\documentclass[letterpaper]{article} 
\usepackage{aaai2026}  
\usepackage{times}  
\usepackage{helvet}  
\usepackage{courier}  
\usepackage[hyphens]{url}  
\usepackage{graphicx} 
\usepackage{natbib}  
\usepackage{caption} 
\usepackage{amsmath}
\usepackage{algorithm}
\usepackage{algpseudocode}
\usepackage{amssymb}
\usepackage{xspace}
\usepackage{booktabs}
\usepackage{multirow}
\usepackage{array}
\usepackage{subcaption}
\usepackage{xcolor}
\usepackage{url}

\usepackage[breakable]{tcolorbox} 
\usepackage{inconsolata}        
\usepackage{newfloat}
\usepackage{listings}
\DeclareCaptionStyle{ruled}{labelfont=normalfont,labelsep=colon,strut=off} 
\floatstyle{ruled}
\newfloat{listing}{tb}{lst}{}
\floatname{listing}{Listing}
\newcommand{\ourname}{{PromptObfus}\xspace}

\title{Anti-adversarial Learning: Desensitizing Prompts for Large Language Models}
\author{
    Xuan Li, Zhe Yin, Xiaodong Gu, Beijun Shen\thanks{The corresponding author.}
}
\affiliations{
    School of Computer Science, Shanghai Jiao Tong University, China\\
    
    \textrm{\{riken01, yin\_zhe, xiaodong.gu, bjshen\}@sjtu.edu.cn}

}

\usepackage{bibentry}

\begin{document}

\maketitle

\begin{abstract}
With the widespread use of LLMs, preserving privacy in user prompts has become crucial, as prompts risk exposing private and sensitive data to cloud LLMs. 
Conventional techniques like homomorphic encryption (HE), secure multi-party computation, and federated learning (FL) are not well-suited to this scenario due to the lack of control over user participation in remote model interactions.
In this paper, we propose \ourname, a novel method for desensitizing LLM prompts. 
The core idea of \ourname is ``anti-adversarial'' learning. Unlike adversarial attacks that add imperceptible perturbations to mislead models, \ourname perturbs sensitive words to make them unrecognizable to humans while maintaining the model's original predictions.
Specifically, \ourname frames prompt desensitization as a masked language modeling task, replacing privacy-sensitive terms with a \texttt{[MASK]} token. A desensitization model is utilized to generate candidate replacements for each masked position. These candidates are subsequently selected based on gradient feedback from a surrogate model, ensuring minimal disruption to task output.
We demonstrate the effectiveness of our approach on three NLP tasks. Results show that \ourname effectively prevents privacy inference from remote LLMs while preserving task utility.
\end{abstract}

\begin{links}
    \link{Code and Datasets}{https://github.com/riken01/PromptObfus}
\end{links}

\section{Introduction}

\label{sec:introduction}
The widespread adoption of large language models (LLMs) such as ChatGPT in various NLP tasks \cite{osti_10510451} has raised significant concerns regarding their inherent privacy risks. Due to the substantial computational resources required for local deployment, users often rely on cloud APIs provided by model vendors, which introduces potential vulnerabilities. Specifically, user-submitted prompts, the primary medium of interaction with LLMs, may inadvertently expose sensitive information, posing serious privacy threats.

\begin{figure}[t!]
\centerline{\includegraphics[width=0.4\textwidth, trim=0 0 0 0 clip]{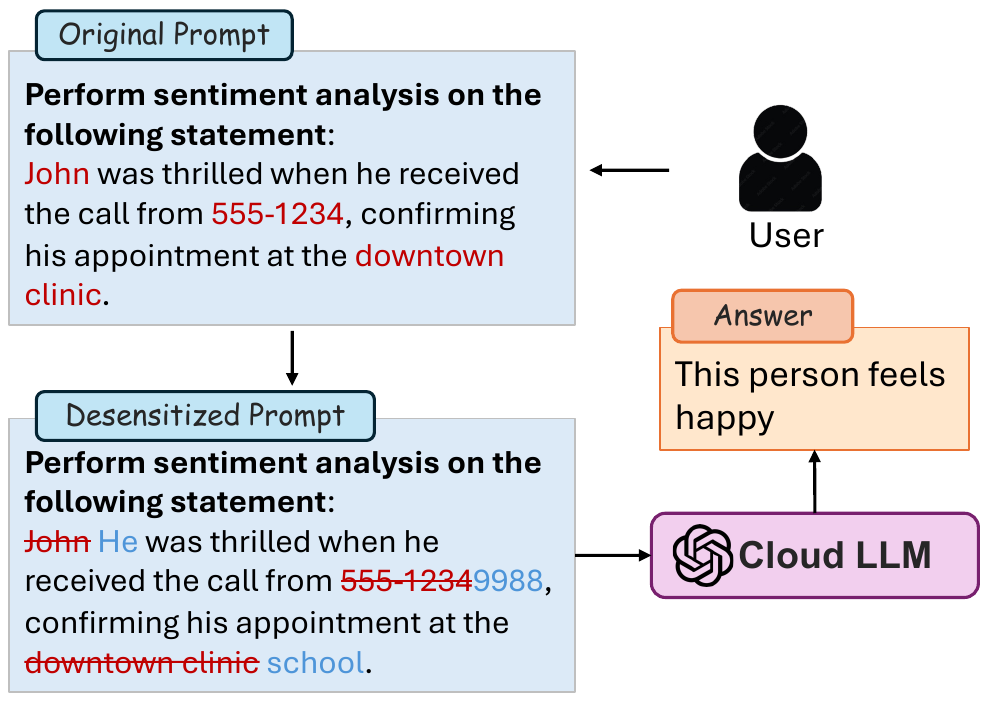}}
    \caption{Illustration of prompt desensitization.} 
    \label{fig:background}
\end{figure}

Prompts often contain personally identifiable information (PII), including names, addresses, and occupational details, as illustrated in Figure \ref{fig:background}. Without proper safeguards during processing, this sensitive data becomes vulnerable to malicious exploitation, leading to serious privacy breaches \cite{duan2024privacyriskincontextlearning}. Thus, developing robust privacy protection mechanisms for LLM prompts has become an urgent research priority.

Conventional privacy-preserving techniques, such as Homomorphic Encryption (HE) \cite{10.5555/1834954}, Secure Multi-Party Computation (MPC) \cite{4568388}, and Federated Learning (FL) \cite{pmlr-v54-mcmahan17a}, exhibit significant limitations when applied to prompts for LLMs, particularly in black-box settings where access to the model's internal architecture or training data is restricted. These methods often fail to simultaneously address the competing requirements of real-time performance, computational efficiency, and robust privacy protection.

Text obfuscation has emerged as a prevalent approach to safeguarding sensitive information in prompts \cite{miranda2025preservingprivacylargelanguage}. For instance, techniques include injecting noise into word embeddings based on differential privacy to perturb sensitive data \cite{yue-etal-2021-differential, gao2024dataadaptivedifferentiallyprivateprompt}, clustering word vectors to render representations of sensitive terms indistinguishable \cite{zhou-etal-2023-textobfuscator}, and training models for data anonymization by detecting and removing PII entities \cite{chen2023hideseekhaslightweight,frikha2025privacyscalpelenhancingllmprivacy}. However, these methods often struggle to achieve an optimal trade-off between privacy preservation and task utility \cite{zhang2024freelunchtheoremprivacypreserving}. Furthermore, approaches that rely on model training typically necessitate expert-annotated datasets, which are challenging to procure in practical applications.

In this paper, we propose \ourname, a portable and task-flexible method for desensitization of LLM prompts. Inspired by the work on generating adversarial examples \cite{alzantot-etal-2018-generating}, we introduce the concept of \textit{anti-adversariality}, which inverts the adversarial objective: rather than crafting subtle perturbations to mislead models, we strategically alter sensitive words to make them unrecognizable to human interpreters while maintaining the model's original task performance. \ourname achieves desensitization by replacing words with semantically distinct yet task-consistent alternatives, thereby ensuring robust privacy protection without compromising the original functionality of the prompts.
\ourname operates through the deployment of two small local models: a \textit{desensitization model}, which replaces sensitive words with privacy-preserving alternatives, and a \textit{surrogate model}, which emulates the task execution of the remote LLM to guide prompt selection. The pipeline consists of three critical steps: generating desensitized alternatives for privacy-sensitive words, assessing the task utility of the LLM, and selecting replacements that minimize performance degradation.


We evaluate \ourname on three NLP tasks: sentiment analysis, topic classification, and question answering. The results demonstrate that our approach establishes new state-of-the-art privacy protection, achieving a 62.70\% reduction in implicit privacy inference attack success rates compared to existing high-accuracy baselines, while completely eliminating explicit inference attacks. Notably, our approach simultaneously preserves competitive task utility, yielding accuracy scores of 86.67\%, 85.25\%, and 96.0\%, respectively.

Our contribution can be summarized as follows:
\begin{itemize}
    \item We introduce the novel concept of \textbf{anti-adversariality}, a pioneering approach for desensitizing LLM prompts that ensures robust privacy protection without compromising task utility.
    
    \item We propose a new privacy-preserving word replacement algorithm, which integrates masked word prediction with LLM gradient surrogation to achieve optimal desensitization.
    
    \item We conduct extensive evaluations of our method across multiple NLP tasks, demonstrating its effectiveness in preserving privacy while preserving task utility.
\end{itemize}

\section{Related Work}
\label{sec:related_work}
\noindent\textbf{Privacy Protection for LLMs.} 
Despite their widespread utility, LLMs raise critical privacy concerns \cite{mireshghallah2024llmssecrettestingprivacy}. Current research addresses these through: (1) model protection via federated learning \cite{10733964, 10.1145/3682068} and homomorphic encryption \cite{NEURIPS2022_64e2449d}; (2) prompt security using encryption \cite{lin2024emojicryptpromptencryptionsecure} and noise-based obfuscation \cite{zhou-etal-2023-textobfuscator,gao2024dataadaptivedifferentiallyprivateprompt}; and (3) PII detection/removal techniques \cite{chen2023hideseekhaslightweight,sun2024depromptdesensitizationevaluationpersonal, chowdhury2025prepsilonepsilonmptsanitizingsensitiveprompts}. Hybrid input strategies mixing real and synthetic data further enhance privacy \cite{utpala-etal-2023-locally}.

\noindent\textbf{Automatic Prompt Engineering.}  
Automatic prompt generation leverages AI to produce privacy-preserving prompts, offering superior performance compared to manual approaches \cite{zhou2022large}. Notable frameworks include APE \cite{yang2024largelanguagemodelsoptimizers}, which iteratively refines prompts by selecting and resampling candidate prompts; APO \cite{zhou2022large}, employing gradient-inspired feedback optimization; and OPRO \cite{pryzant-etal-2023-automatic}, utilizing LLMs as meta-optimizers for prompt improvement.

\noindent\textbf{Text Adversary Generation.}  
Adversarial training is a technique aimed at improving model robustness against malicious or deceptive inputs, widely applied in domains such as computer vision, NLP, and speech recognition. In this approach, models are systematically exposed to adversarial examples \cite{10.5555/2969033.2969125}, which are inputs subtly modified to induce significant changes in model outputs. Genetic algorithms are employed to generate semantically equivalent adversarial samples \cite{alzantot-etal-2018-generating}, selecting synonyms that maximize the likelihood of the target label. More recently, LLMs are utilized to produce adversarial samples \cite{wang2023generatingvalidnaturaladversarial}.

In contrast to existing approaches, we propose an \textit{anti-adversarial} method for the desensitization of LLM prompts, which ensures that model outputs remain consistent while rendering sensitive content imperceptible to human interpretation.

\section{Methodology}
\label{sec:methodology}

\begin{figure*}[t!]
\centerline{\includegraphics[width=0.75\textwidth, trim=0 -3mm 0 8mm clip]{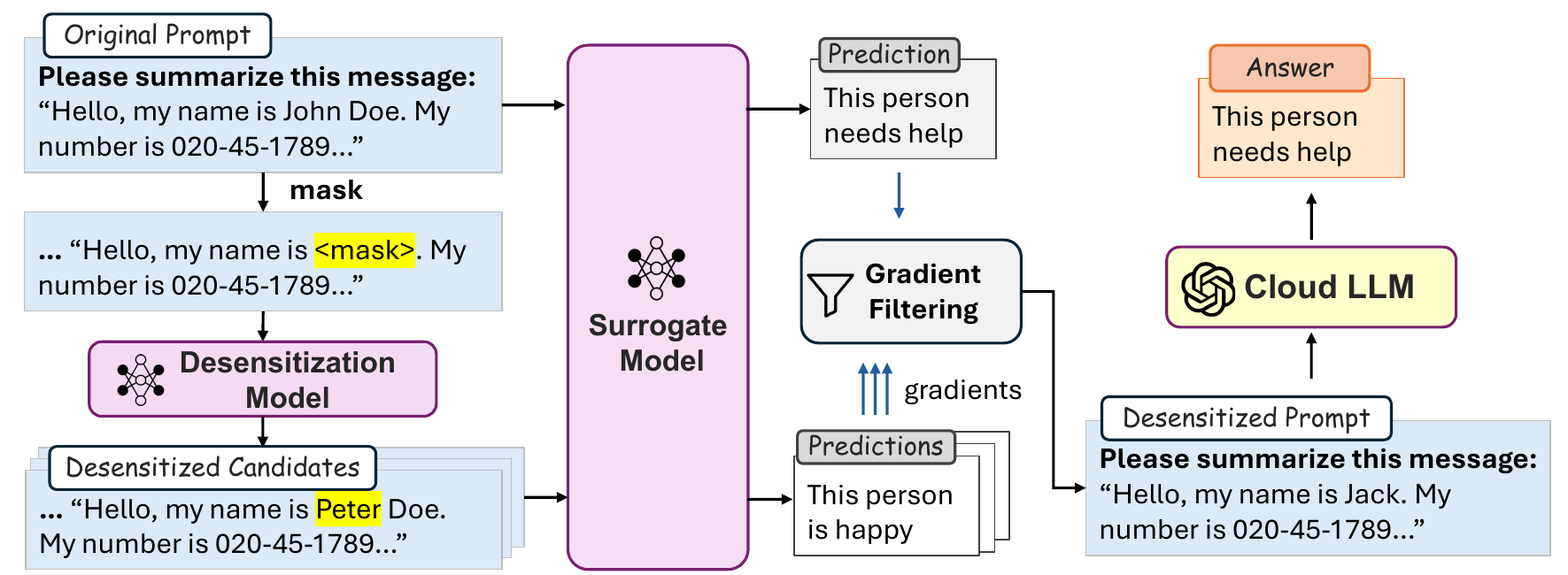}}
    \caption{Overview of \ourname.} 
    \label{fig:overview}
\end{figure*}

Inspired by the principles of adversarial example generation \cite{alzantot-etal-2018-generating}, we conceptualize our approach as an \textit{anti-adversarial} framework, wherein the objective is to obfuscate sensitive information while preserving the original behavior and predictive performance of the model.

\subsection{Problem Statement}
Consider an LLM \( \Phi(y|x) \) with parameters \( \Phi \) and a downstream task (\emph{e.g.}, question answering) characterized by a parallel dataset \( \mathcal{T} = \{(x^{(i)}, y^{(i)})\}_{i=1}^N \), where \( x \) and \( y \) represent the input prompt and target output, respectively. We formulate the following privacy-preserving transformation problem: Given a set of privacy attributes \( P = [p_1, \dots, p_m] \) and an input \( x = \{x_1, \dots, x_n\} \), our goal is to derive a desensitized prompt \( x' = \{x'_1, \dots, x'_n\} \) that eliminates all \( P \)-attributes while preserving task utility. Formally:
\begin{equation}
  \begin{split}
    & \min_{x'=M(x|\lambda,k)}  \|s(\Phi(x'), y) - s(\Phi(x), y)\| \\
    & \ \ \ \ \ s.t. \ \ \  x'_i\notin P \ \ \ \forall x'_i\in x'\\
  \end{split}
    \label{eq:optimization}
\end{equation}
where \( M(x|\lambda, k) \) denotes a desensitization mapping function, \( \lambda \) controls the candidate replacement set size for each sensitive term, and \( k \) modulates the confusion ratio. The task-specific metric \( s: Y \times Y \rightarrow \mathbb{R} \) (\emph{e.g.}, BLEU for QA) evaluates utility preservation.

\subsection{Overview}
Our approach is designed to optimize the desensitization function \( M(x|\lambda,k) \) to preserve LLM output fidelity while eliminating privacy risks.  
Figure \ref{fig:overview} illustrates the overall architecture of \ourname. 
The pipeline consists of three steps: (1) detecting privacy attributes and generating candidate replacements using a dedicated desensitization model; (2) assessing utility preservation through a surrogate model by comparing with the original prompt's performance; and (3) performing gradient-based optimization to select the most suitable replacements from candidates, ultimately producing the final privacy-preserving prompt.

\subsection{Predicting Candidate Desensitive Words}

For each privacy-sensitive word in an input prompt, \ourname generates a set of candidate replacements through desensitization. This process can be formalized as a Masked Language Model (MLM) task, where privacy-sensitive words are substituted with a \texttt{mask} token. The desensitization model is utilized to predict precisely \(\lambda\) candidate desensitized replacements for each masked position. By leveraging pre-trained semantic representations, the model ensures all candidate replacements maintain contextual appropriateness relative to the surrounding text. This approach preserves textual coherence and prompt functionality while effectively concealing sensitive information through semantically valid substitutions.

We utilize spaCy's NER model~\cite{NER} to detect explicit privacy attributes like person names, locations, and organizations. 
All identified privacy-sensitive words are uniformly replaced with \textsc{MASK} tokens. 
Beyond explicit attributes, we address potential implicit privacy risks through contextual analysis. Specifically, we mask rare words identified by their TF-IDF scores \cite{10448234,10.5555/106765.106782}, as these terms are statistically more likely to contain identifiable information. The top $k$ highest-scoring terms are selected for masking.


\begin{figure}[t!]
\centerline{\includegraphics[width=0.4\textwidth, trim=0 0 0 -5mm clip]{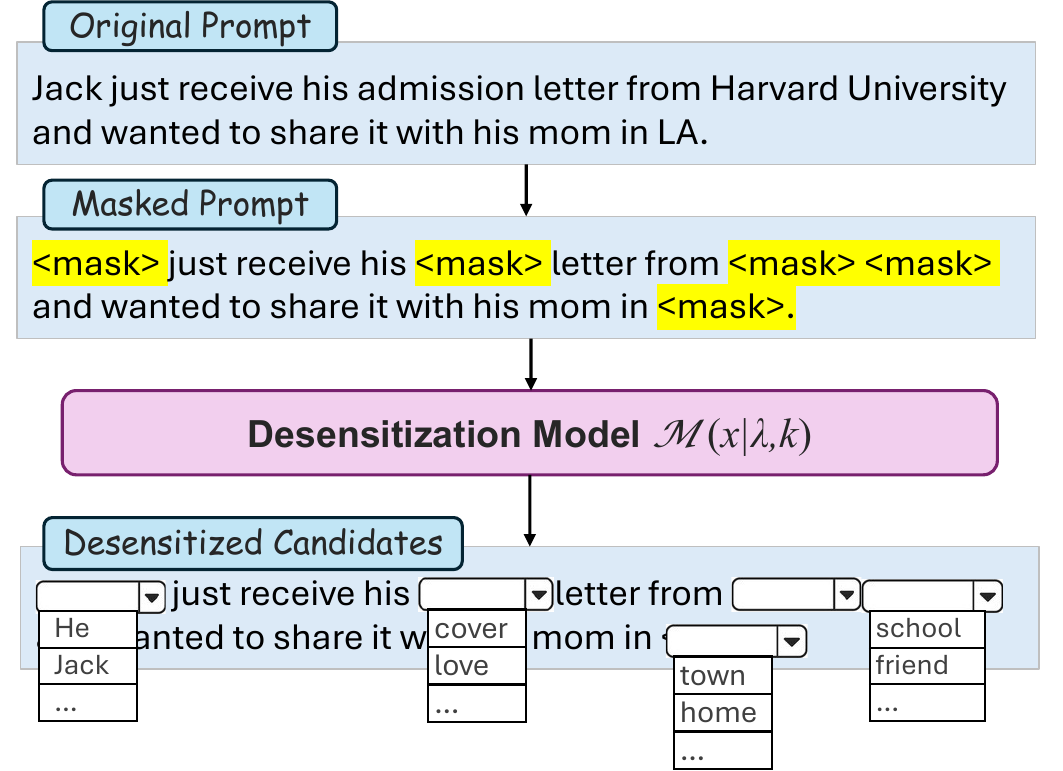}}
    \caption{Predicting candidate desensitive words.} 
    \label{fig:example-original}
\end{figure}

Next, a pre-trained language model, referred to as the \textit{desensitization model}, is utilized to generate potential replacement candidates for each masked token, as shown in Figure \ref{fig:example-original}. This model can employ any pre-trained language architecture with MLM capability, such as RoBERTa.

To mitigate the risk of privacy leakage through synonyms or near-synonyms, the desensitized word set is further refined by assessing semantic similarity. 
This is achieved by computing the Euclidean distance between the word embeddings of each candidate \( w_i \) and the original word \( x_{\text{original}} \):
\begin{equation}
   d(x_{\text{original}}, w_i) = \|\vec{x_{\text{original}}} - \vec{w_i}\|
\end{equation}
where \( \vec{x_{\text{original}}} \) and \( \vec{w_i} \) are the word vectors, and \( \|\cdot\| \) is the Euclidean norm. To this end, we apply an empirically determined distance threshold \( \theta_{\text{dist}} = 0.95 \). All candidates satisfying \( d(x_{\text{original}}, w_i) \leq \theta_{\text{dist}} \) are considered semantically too similar and are consequently removed. 
The resulting filtered set \( W_{\text{filtered}} \) is defined as:
\begin{equation}
   W_{\text{filtered}} = \{ w_i \in W \mid d(x_{\text{original}}, w_i) > \theta_{\text{dist}} \}
\end{equation}

\subsection{Assessing Task Utility}

To preserve task utility, we design a gradient-based selection criterion for desensitized words. 
Gradient magnitudes serve as indicators of input sensitivity: larger values suggest substantial semantic distortion from word replacement, while smaller values imply better semantic preservation with minimal output perturbation.

Since direct gradient acquisition from remote LLMs is infeasible, \ourname employs a smaller white-box surrogate model \( \mathcal{M}_{surrogate} \) to approximate the target LLM's behavior, which captures how variations in the prompt affect the performance signal. This computationally efficient alternative enables both task evaluation and gradient computation while maintaining manageable resource requirements. \ourname supports two types of surrogate models: 

1) {Task-specific model:} When adequate task-specific data \( \mathcal{D} = \{(x, y)\} \) exists, a lightweight fine-tuned model provides precise, task-aware gradient estimates for prompt desensitization.

2) {General model:} For data-scarce scenarios, a moderately-sized pre-trained language model (still substantially smaller than target LLMs)  serves as the surrogate. This variant produces less task-specific but more generalizable gradient approximations.

\subsection{Gradient Filtering}
\ourname utilizes gradient magnitudes from the surrogate model \( \mathcal{M}_{surrogate} \) to assess desensitized candidates in \( W_{filtered} \), selecting the word corresponding to the minimal gradient value.


For each candidate word \( w \in W_{filtered} \), \ourname generates a modified prompt \( x' \) and computes its output gradient. Formally, the gradient magnitude is calculated as:
\begin{equation}
    \begin{split}
     \Delta_i(w) = \left\| \frac{\partial \mathcal{L}(y, \mathcal{M}_{surrogate}(x'[i \leftarrow w]))}{\partial x'} \right\|   
    \end{split}
\end{equation}
where \( i \) indicates the target word position, \( \Delta_i(w) \) captures the gradient sensitivity, and \( \mathcal{L} \) represents the task loss function. Through iterative evaluation, the optimal replacement \( w^* \) is selected via:
\begin{equation}
    \begin{split}
w^* = \arg\min_{w \in W_{filtered}} \Delta_i(w)
    \end{split}
\end{equation}

Finally, \ourname substitutes the privacy-sensitive word at position \( i \) with the optimal replacement \( w^* \), iterating this procedure across all masked positions. This sequential filling approach selects each replacement by considering both local contextual constraints and global semantic coherence from prior substitutions, thereby preserving task utility while preserving text semantics.

\section{Experiments Setup}

We evaluate the effectiveness of \ourname across two critical dimensions, emphasizing its capacity to maintain robust privacy protection while preserving task utility. To demonstrate its practical utility, we apply \ourname to three NLP tasks: sentiment analysis, topic classification, and question answering. These tasks represent diverse real-world applications and provide a comprehensive assessment of the method's applicability.

To evaluate \ourname's privacy protection capabilities, we simulate adversarial attacks to assess whether sensitive information can be extracted from desensitized prompts. We consider three attack strategies, including two text reconstruction methods and one privacy inference method: Embedding Inference (EI), Mask Token Inference (MTI), and PII Inference.
\textit{EI} \cite{10.1145/3459637.3482281} measures the semantic similarity between each word representation and a publicly available word embedding matrix, predicting sensitive content based on the nearest neighbors.
\textit{MTI} \cite{yue-etal-2021-differential} masks tokens in desensitized prompts and assesses the attacker's success in reconstructing the original text.
\textit{PII Inference} \cite{plant-etal-2021-cape} examines textual patterns to deduce private user attributes.

\subsection{Baselines}
We compare \ourname against six state-of-the-art privacy-preserving methods and the original unprotected text. 
1) \textbf{Random Perturbation}, which randomly substitutes a portion of tokens in the text with arbitrary words. 
2) \textbf{Presidio} \cite{Presidio}, an automated tool for detecting and redacting sensitive information, including names, locations, and other personally identifiable information.
3) \textbf{SANTEXT} \cite{yue-etal-2021-differential}, a differential privacy approach that determines word replacement probabilities based on Euclidean distances in embedding space.
4) \textbf{SANTEXT+} \cite{yue-etal-2021-differential}, an improved variant of SANTEXT that incorporates word frequency information to optimize replacement probabilities.
5) \textbf{DP Prompt} \cite{utpala-etal-2023-locally}, a method that employs LLMs to paraphrase original prompts while preserving privacy.
6) \textbf{PromptCrypt} \cite{lin2024emojicryptpromptencryptionsecure}, which transforms original prompts into emoji sequences using large models.

\subsection{Evaluation Metrics}

\noindent\textbf{Privacy Protection Metrics}.
We measure the potential leakage of private information to third-party attackers through quantitative evaluation. Two key metrics are adopted to assess privacy protection performance: \textit{TopK Accuracy} and \textit{Success Rate}.
\textit{TopK Accuracy} \cite{zhou-etal-2023-textobfuscator} evaluates token-level privacy by computing the proportion of correctly inferred words among the top \textit{k} predictions generated by third-party attackers. \textit{Success Rate} \cite{plant-etal-2021-cape} measures the exposure risk of personally identifiable information by determining the percentage of successfully extracted PII entities relative to the total identifiable information present.

\noindent\textbf{Task Utility Metrics}.
To assess \ourname's capability in preserving task utility, we measure the model's accuracy when processing desensitized prompts. Our evaluation employs two standard metrics: accuracy and answer quality score.
\textit{Accuracy} quantifies the proportion of correct predictions relative to the total number of test instances, applicable to both classification and question answering tasks. 
\textit{Answer Quality Score} evaluates the overall quality of responses, considering factors including correctness, relevance, completeness, and readability.
For automated assessment, we employ GPT-4o-mini as an evaluator, with the complete scoring rubric provided in Appendix \ref{sec:prompt}.

\subsection{Datasets}

\begin{table}[t]
\centering
\small
\begin{tabular}{c|c|c}
\toprule
\textbf{Dataset} & \textbf{Split} & \textbf{Number of Samples} \\ \midrule
\multirow{3}{*}{\textbf{SST-2}} & Train & 67,349 \\ \cline{2-3} 
 & Validation & 872 \\ \cline{2-3} 
 & Test & 1,821 \\ \midrule
\multirow{3}{*}{\textbf{AG News}} & Train & 120,000 \\ \cline{2-3} 
 & Validation & 7,600 \\ \cline{2-3} 
 & Test & 7,600 \\ \midrule
 \textbf{PersonalPortrait} & Test & 400 \\ \bottomrule
\end{tabular}
\caption{Statistics of the datasets.} 
\label{tab:dataset_distribution}
\end{table}

\begin{table*}[t!]
\small
\centering
\begin{tabular}{lccccc}
\toprule
\textbf{Approach}    & \textbf{Acc.\( \uparrow \)} & \textbf{MTI Top1\( \downarrow \)} & \textbf{EI Top1\( \downarrow \)} & \textbf{PI Success Rate\( \downarrow \)} & \textbf{Avg. Ranking\( \downarrow \)} \\ \midrule
Origin            & 87.50               & 31.37             & --            & --                  & --  \\
\midrule
Random          & 83.75 (4)             & 17.10 (2)             & 83.78 (7)        & 97.50 (9)      &   5.50         \\
Presidio & 83.25 (6)               & 23.28 (5)             & 71.53 (6)        & \textbf{0.00 (1)}   &   4.50                 \\
SANTEXT         & 61.50 (8)              & 21.43 (3)             & 62.10 (5)        & 41.75 (7)     &   5.75           \\
SANTEXT+        & 55.25 (9)              & \textbf{11.04 (1)}             & \textbf{49.09 (1)}        & 34.25 (6)       &   4.25          \\ 
DP-Prompt         & 85.00 (2)              & --             & --        & 96.25 (8)     &    5.00\\
PromptCrypt         & 72.00 (7)              & --             & --        & 13.50 (5)   &   6.00              \\ 
\midrule
\ourname (k=0.1)  & \textbf{85.25 (1)}              & 24.68 (7)             & 61.86 (4)        & \textbf{0.00 (1)}      & 3.25                \\
\ourname (k=0.2)  & 84.50 (3)              & 23.31 (6)             & 55.82 (3)        & \textbf{0.00 (1)}     & 3.25             \\
\ourname (k=0.3)  & 83.75 (4)              & 22.89 (4)             & 49.65 (2)        & \textbf{0.00 (1)}   & \textbf{2.75}               \\
\bottomrule
\end{tabular}
\caption{Performance of privacy protection and task utility with detailed rankings on the AG News topic classification task. In the PI Attack, the AG News dataset does not explicitly label privacy attributes. Therefore, the attack assumes that named entities (\emph{e.g.}, person names, locations) represent explicit privacy attributes and targets these for evaluation. The individual rankings are indicated in ( ). } 
\label{tab:topic-results}
\end{table*}

\begin{table*}[t]
\small
\centering
\begin{tabular}{lccccccc}
\toprule
\textbf{Approach}    & \textbf{Acc.\( \uparrow \)} & \textbf{Quality Score\( \uparrow \)} & \textbf{MTI Top1\( \downarrow \)} & \textbf{EI Top1\( \downarrow \)} & \textbf{PI(Loc.)\( \downarrow \)} & \textbf{PI(Occ.)\( \downarrow \)} & \textbf{Avg. Ranking\( \downarrow \)} \\ \midrule
Origin         & 96.9               & 3.86  & 46.43 &  --              & 94.75                       & 60.25         & --                \\
\midrule
Random         & 90.0 (8)               & 3.34 (6)          & \textbf{32.67 (1)} &   90.00 (6)     & 81.50 (8)                      & 46.25 (5)      & 5.67                    \\
Presidio         & \textbf{96.9 (1)}               & 3.56 (4)         & 44.16 (5) &    96.62 (7)    & \textbf{0.00 (1)}                        & 55.00 (8)   & 4.33
\\ 
SANTEXT         & 91.0 (6)               & 3.27 (8)     & 55.75 (6) &     78.56 (4)       & \textbf{0.00 (1)}                        & 47.00 (6)        & 5.17                 \\
SANTEXT+         & 91.3 (5)               & 3.33 (7)   & 55.75 (6) &   \textbf{61.62 (1)}            & \textbf{0.00 (1)}                        & 48.25 (7)       & 4.50                  \\
DP-Prompt         & 95.0 (3)               & 3.62 (2)   & --  &  --             & 89.25 (9)                       & 55.25 (9)        & 5.75                 \\
PromptCrypt         & 49.5 (9)               & 2.89 (9) & --  & --               & 16.25 (7)                        & \textbf{11.00 (1)}            &6.50             \\
\midrule
\ourname (k=0.1)              & 96.0 (2)               & \textbf{3.63 (1)}   & 42.30 (4) &  86.45 (5)            & \textbf{0.00 (1)}                      & 45.75 (4)          & \textbf{2.83}                \\
\ourname (k=0.2)              & 93.0 (4)              & 3.61 (3)           & 38.81 (3)  &   77.02 (3)   & \textbf{0.00 (1)}                      & 37.75 (3)         &    \textbf{2.83}              \\
\ourname (k=0.3)              & 90.5 (7)               & 3.46 (5)        & 36.57 (2) &    68.10 (2)     & \textbf{0.00 (1)}                      & 17.25 (2)         & 3.16                 \\
\bottomrule
\end{tabular}
\caption{Performance of privacy protection and task utility with detailed rankings on the PersonalPortrait text QA task. In the PI Attack, we annotate two types of private entities in the data: Location and Occupation. Among them, Occupation is considered implicit privacy, as it is more likely to be inferred from the context. The individual rankings are indicated in ( ).} 
\label{tab:QA-results}
\end{table*}

Our evaluation employs two established benchmark datasets: \textbf{SST-2} \cite{socher-etal-2013-recursive} for sentiment analysis and \textbf{AG News} \cite{NIPS2015_250cf8b5} for topic classification. Since existing QA datasets typically contain anonymized or desensitized content and are therefore unsuitable for privacy evaluation, we develop \textbf{PersonalPortrait}, a specialized dataset comprising 400 sensitive psychological counseling dialogues. These patient narratives are generated using GPT-4 and subsequently validated through rigorous manual review by two domain experts to ensure both authenticity and privacy relevance. Complete dataset statistics are presented in Table \ref{tab:dataset_distribution}, while the detailed construction process is documented in Appendix \ref{sec:PersonalPortrait}.

\subsection{Implementation Details}
We implement \ourname by utilizing three open-source language models: {RoBERTa-base}~\cite{RoBERTa-base} as the core desensitization model, {BART-large}~\cite{BART-large} as the task-specific surrogate model for classification tasks, and {GPT-Neo-1.3B}~\cite{GPT-Neo-1.3B} as the general surrogate model for question answering tasks, selected based on dataset size considerations. 

To ensure a fair comparison, we maintain a consistent obfuscation ratio across all word-level protection baselines and \ourname. 
As DP Prompt and PromptCrypt operate at the prompt level rather than the word level, they are evaluated solely using PI Attack rather than MTI or EI Attack. All experiments employ the original parameter configurations from their respective publications, with GPT-4o-mini implemented as the remote LLM.
Further details on hyperparameter configurations are provided in Appendix \ref{sec:trainingsettings}.

\section{Results and Analysis}

\subsection{Overall Performance}
Tables \ref{tab:topic-results}  and \ref{tab:QA-results} present the experimental results on the AG News and PersonalPortrait datasets, respectively (further details for other datasets are provided in Appendix \ref{sec:otherresults}). \ourname demonstrates superior performance with an average ranking of 2.75 and 2.83, respectively, surpassing all baseline methods.

\smallskip\noindent\textbf{Privacy Protection}. The PI attack of explicit privacy attains a 0.00\% success rate against \ourname-generated prompts, confirming complete explicit privacy preservation. Comparative methods (SANTEXT+, DP Prompt, PromptCrypt) exhibit significantly higher vulnerability, as they modify linguistic structures rather than implementing targeted PII protection. In the PI Inference of \textit{Occupation}, \ourname achieves the second-lowest attack success rate at 17.25\%, trailing only PromptCrypt (11.00\%). Compared against high-accuracy baselines exceeding 90\% accuracy, \ourname attains a 62.70\% decrease in implicit privacy inference attack success rates. 

\smallskip\noindent\textbf{Task Utility Preservation.} In the topic classification task (AG News, Table \ref{tab:topic-results}), \ourname maintains 85.25\% classification accuracy at $k=0.1$, only a 2.57\% decrease from the original text. This performance exceeds that of other word-level protection techniques, such as Presidio (83.25\%) and SANTEXT+ (55.25\%). In the PersonalPortrait dataset (QA task, Table \ref{tab:QA-results}), \ourname achieves 96.0\% accuracy, closely matching the original text's performance (96.9\%) with merely 0.93\% degradation. 

These results collectively indicate that \ourname successfully achieves robust privacy protection against remote LLM attacks while preserving the original model's task performance, establishing an optimal privacy-utility tradeoff among all evaluated methods.

\subsection{Ablation Studies}

\noindent\textbf{Impact of Surrogate Model.}
We investigate the impact of architectures and scales of the surrogate model across three model types: encoder-only (RoBERTa), decoder-only (GPT2), and encoder-decoder (BART) architectures. The evaluation spans three model sizes: base ($\sim$130M parameters, \emph{e.g.}, RoBERTa-base), medium ($\sim$350M parameters, \emph{e.g.}, RoBERTa-large, BART-large, GPT-2-medium), and large (Llama-2-7B, ChatGLM3-6B). Limited by computational resources, we employ full-parameter fine-tuning for small and medium models, while utilizing Low-Rank Adaptation (LoRA) for large models.


\begin{table}[t]
\small
\centering
\begin{tabular}{lccc}
\toprule
\textbf{Approach} & \textbf{Acc.\( \uparrow \)} & \textbf{MTI Top1\( \downarrow \)} & \textbf{EI Top1\( \downarrow \)}  \\
\midrule
Original Data & 87.20 & 48.86 & --  \\
\midrule
GPT2-base & 84.00 & 42.34 & 81.80  \\
Roberta-base & 84.80 & 42.66 & 81.71  \\
BART-base  & 86.40 & 42.47 & 81.73 \\
\hline
GPT2-medium & 84.53 & 43.99 & 81.75  \\
Roberta-large & 85.87 & 42.51 & 81.71  \\
BART-large & \textbf{86.67} & 42.94 & 81.72  \\
\hline
Llama-2-7B & 84.80 & 42.47 & 81.75  \\
ChatGLM3-6B  & 84.53 & 42.94 & 81.72  \\
\bottomrule
\end{tabular}
\caption{Impact of surrogate model variations on obfuscation effectiveness in sentiment analysis.} 
\label{tab:sentiment-RQ2}
\vspace{-8pt}
\end{table}

\begin{table}[t!]
\small
\centering
\begin{tabular}{lccc}
\toprule
\textbf{Approach (k=0.1)} & \textbf{Acc.\( \uparrow \)} & \textbf{MTI Top1\( \downarrow \)} & \textbf{EI Top1\( \downarrow \)} \\
\midrule
\ourname & \textbf{85.25} & 24.68 & \textbf{61.86} \\
\midrule
Random masking & 84.25 & \textbf{24.33} & 63.98 \\

\bottomrule
\end{tabular}
\caption{Performance of privacy protection and task utility on the AGNews topic classification task evaluated under random masking and \ourname.} 
\label{tab:masking-strategy}
\end{table}

The experimental results for sentiment analysis are presented in Table \ref{tab:sentiment-RQ2}, with corresponding question answering results provided in Appendix \ref{sec:surmodel-qa}. We observe that privacy protection efficacy remains unaffected by either the architecture or scale of the surrogate model. Medium-scale models demonstrate superior performance compared to their larger counterparts, as the task complexity does not warrant additional model capacity, and LoRA may limit fine-tuning effectiveness. Encoder-decoder architectures achieve optimal performance by effectively integrating the encoder's classification capabilities with the decoder's alignment to remote model requirements.

\smallskip\noindent\textbf{Impact of Masking Strategy.}
We examine the effectiveness of different masking strategies in preventing implicit privacy leakage. Our comparison focuses on two approaches: random masking, where tokens are selected uniformly at random, and \ourname, a TF-IDF-based method that targets the least frequent tokens. Experimental results on the AGNews dataset reveal that \ourname achieves superior performance in both privacy protection and utility preservation compared to random masking, as detailed in Table \ref{tab:masking-strategy}.

\begin{table}[t!]
\small
\centering
\begin{tabular}{lccc}
\toprule
\textbf{Approach (k=0.1)} & \textbf{Acc.\( \uparrow \)} & \textbf{MTI Top1\( \downarrow \)} & \textbf{EI Top1\( \downarrow \)} \\
\midrule
\ourname & \textbf{85.25} & \textbf{24.68} & \textbf{61.86} \\
\midrule
Top-1 Selection & 84.75 & 35.06 & 79.50 \\
Random Selection & 83.75 & 25.14 & 66.52 \\
$<$MASK$>$ & 83.25 & 27.55 & 63.96 \\
\bottomrule
\end{tabular}
\caption{Performance of privacy protection and task utility on the AGNews topic classification task evaluated under different strategies for selecting the candidate desensitized words.} 
\label{tab:gradient-filtering}
\vspace{-8pt}
\end{table}

\begin{figure}[t!]
\centerline{\includegraphics[width=0.25\textwidth]{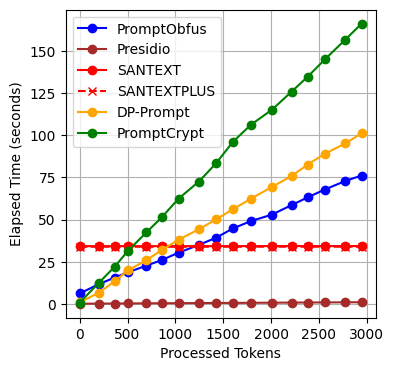}}
    \caption{Elapsed time increases linearly with the number of processed tokens across different methods.} 
    \label{fig:time-efficiency}
\end{figure}

\begin{table*}[!h]
\centering
\small
\renewcommand{\arraystretch}{1.1} 
\begin{tabular}{lp{12.3cm}}
\toprule
\textbf{Original Text}: & I’m a \colorbox{red!30}{39}-year-old \colorbox{orange!30}{driver} in \colorbox{red!30}{Toronto}, and I often feel like my emotions are all over the place... \\ \hline \hline

\bf Random: & abuser a \colorbox{red!30}{39}-year-old \colorbox{orange!30}{driver} in \colorbox{red!30}{Toronto}, moha palmery often feel like my emotions are all over shady place... \\ 
\bf Presidio: & I’m a $<$DATE$>$ \colorbox{orange!30}{driver} in $<$GPE$>$, and I often feel like my emotions are all over the place... \\ 
\bf SANTEXT: & jagger rehashed a hardy - year - old \colorbox{orange!30}{driver} in women , and obscure often feel like my emotions are all over the place...  \\ 
\bf SANTEXT+: & jagger rehashed a fidel 15 year 3 old \colorbox{orange!30}{driver} in motion , and esoteric seldom feel like my emotions are all putting the however...  \\
\bf DP-Prompt: & I’m a \colorbox{red!30}{39}-year-old \colorbox{orange!30}{driver} in \colorbox{red!30}{Toronto}, and my emotions can be unpredictable...\\ 
\bf PromptCrypt: &  \colorbox{red!30}{39} \includegraphics[width=0.35\textwidth]{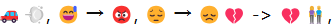}...
\\
\midrule
\bf \ourname (k=0.1): & I’m a commercial \colorbox{orange!30}{driver} of two and I often feel like my emotions are all over the place... \\ 
\bf \ourname (k=0.2): & I’m a commercial assistant in LA and I often feel like my emotions flow all over the world... \\ 
\bf \ourname (k=0.3): & I’m one professional assistant in general and I often feel like my emotions are hovering throughout... \\ 
 \bottomrule
\end{tabular}
\caption{A case of desensitized prompts generated by various methods for question answering.} 
\label{tab:example2}
\vspace{-6pt}
\end{table*}

\begin{table}[t!]
\small
\centering
\resizebox{\columnwidth}{!}{%
\begin{tabular}{cccc}
\toprule
\textbf{Model} & \textbf{GPT-4o-mini} & \textbf{GLM-4-plus} & \textbf{Meta AI} \\
\midrule
GPT2 & 84.53 & 91.2 & \textbf{91} \\

ChatGLM3-6B & 84.53 & 91.0 & 89 \\

Llama2-7B & 84.80 & 90.8 & 90 \\
\midrule

BART & \textbf{86.67} & \textbf{91.4} & \textbf{91} \\
\bottomrule
\end{tabular}
}
\caption{Classification accuracy of local-remote model combinations on the sentiment analysis (SST) task. Columns denote remote models, while rows denote local models.} 
\label{tab:RQ3}
\end{table}

\smallskip\noindent\textbf{Impact of Gradient Filtering.}
We evaluate four candidate selection strategies for word desensitization: (1) \ourname's default gradient-based strategy, which minimizes downstream task impact by selecting tokens with the smallest output gradient magnitudes while excluding original words and their synonyms through additional filtering; (2) top-1 prediction based on model confidence; (3) random selection from candidates; and (4) direct `$<$MASK$>$' token insertion as a baseline. We set the number of candidate desensitized words ($\lambda$) to 10. 
The experimental results on the AGNews dataset are shown in Table \ref{tab:gradient-filtering}, which demonstrate that \ourname's gradient-based approach achieves an optimal balance between privacy preservation and task utility. 
 
Additionally, a detailed examination of hyperparameters \( k \) and \( \lambda \) is presented in Appendix \ref{sec:hyperparameters}.

\subsection{Time Efficiency Evaluation}
We evaluate \ourname's computational efficiency on an NVIDIA RTX 3090 GPU with CUDA v12.4. All comparative methods are executed under identical configurations to ensure fair comparison.
The results are presented in Figure \ref{fig:time-efficiency}. Our method achieves an optimal balance between computational efficiency and privacy preservation. Notably, the system exhibits a processing rate of 100 tokens in 2.58 seconds, demonstrating practical runtime performance for real-world applications.

\subsection{Transferability}
We further explore the transferability of trained surrogate models across different platform combinations. Experiments evaluate local-remote model pairings from three providers: OpenAI, Meta, and Zhipu.
Experimental results, presented in Table \ref{tab:RQ3}, indicate that cross-platform model combinations maintain comparable obfuscation effectiveness, showing strong transferability across vendors. 
For additional validation, we test BART-large, the best-performing independent model from previous experiments, with all three remote models. The results consistently show BART-large's superior performance in every configuration.

\subsection{Case Study}

Table \ref{tab:example2} illustrates an example of desensitized prompts generated by various methods for question-answering. The original text contains identifiable sensitive information, including age (`39-year-old'), occupation (`driver'), and location (`Toronto'). \ourname successfully replaces explicit private attributes (age, location) with de-identified terms, ensuring robust privacy protection. At \( k=0.2 \) and \( k=0.3 \), the obfuscation intensity increases, and implicit privacy details, such as occupation (`driver'), are substituted with more ambiguous terms like `assistant' while preserving semantic coherence and readability.

In contrast, the Random method fails to accurately identify and modify sensitive information, leading to the leakage of all privacy-related terms and a lack of textual coherence. Presidio is limited to handling predefined temporal and geographic patterns, offering insufficient flexibility and failing to protect occupation-related privacy. Meanwhile, SANTEXT and SANTEXT+ introduce excessive noise, rendering the sentences overly disordered and degrading task utility. DP-Prompt results in privacy leakage, while PromptCrypt, though privacy-preserving, employs overly simplistic and abstract symbols, causing significant performance degradation.

\section{Conclusion}
\label{sec:conclusion}

In this paper, we introduce \ourname, a novel method for privacy-preserving prompt desensitization in LLMs.
Its core idea is \textit{anti-adversarial learning}, which simultaneously preserves model output fidelity while preventing human interpretation of sensitive content. \ourname achieves this by replacing sensitive words in user prompts with semantically distant yet task-consistent alternatives, minimizing the impact on task utility. 
Evaluations across three NLP tasks demonstrate \ourname's effectiveness in safeguarding privacy against cloud-based LLM attacks while maintaining original task utility levels. The results establish \ourname's superior privacy-utility balance compared to existing baseline methods.

\section*{Acknowledgments}
This research is funded by the National Key Research and Development Program of China (Grant No. 2023YFB4503802), the National Natural Science Foundation of China (Grant No. 62032004), and the Natural Science Foundation of Shanghai (Grant No. 25ZR1401175).

\bibliography{main}
\clearpage

\setcounter{secnumdepth}{2} 

\appendix

\appendix

\section{Appendix}
\label{sec:appendix}

\subsection{PersonalPortrait Construction}
\label{sec:PersonalPortrait}

Inspired by the D4 dataset \cite{yao-etal-2022-d4} and the PersonalReddit dataset \cite{Staab2023BeyondMV}, which synthesize text from personal profiles, we develop authentic patient profiles incorporating demographic characteristics (gender, occupation, location) and psychiatric conditions to simulate clinical counseling dialogues. The QA task focuses on mental health diagnosis, where models must identify specific disorders (\emph{e.g.}, depression, anxiety) through symptom analysis (\emph{e.g.}, insomnia, persistent sadness, anxious thoughts) present in the counseling transcripts.

The dataset synthesis involves three key stages:
\begin{enumerate}
    \item {Profile generation}: Diverse patient profiles are created by randomly sampling demographic attributes (age, gender, location, occupation) and psychiatric conditions from predefined clinical categories.
    
    \item {Text generation}: GPT-4 synthesizes clinically coherent self-reports based on generated profiles, capturing authentic symptom presentations and personal histories (see Appendix \ref{sec:prompt} for generation templates).
    
    \item {Clinical validation}: Generated texts undergo expert review to verify diagnostic accuracy, symptom plausibility, and therapeutic appropriateness.
\end{enumerate}

Each entry in the PersonalPortrait dataset contains:
\begin{itemize}
\item {Clinical Narrative}: A first-person account simulating psychological counseling sessions, documenting symptom presentation, behavioral patterns, and psychosocial experiences related to mental health conditions.
\item {Age:} Integer values uniformly distributed between 18 and 65 years.
\item {Location:} Randomly sampled from 20 predefined English-speaking geographical regions.
\item {Gender:} Binary classification (male/female) with equal probability.
\item {Occupation:} Profession randomly selected from a standardized list of 20 common occupations.
\item {Clinical Diagnosis:} Mental health disorder randomly assigned from 10 prevalent psychiatric conditions following DSM-5 classification criteria.
\end{itemize}

\subsection{Prompt Templates}
\label{sec:prompt}

We present the core prompt templates utilized in our study, where \textit{TEXT} indicates the desensitized prompt, \textit{ORIGINAL TEXT} corresponds to the unprotected original prompt, and \textit{ANSWER} signifies the LLM-generated response. Attributes such as \textit{AGE} and \textit{GENDER} are randomly sampled from predefined categories.

\begin{tcolorbox}[colframe=black, colback=white, coltitle=white, colbacktitle=black, 
title=Prompt for Generating Samples for PersonalPortrait,
boxrule=0.8pt,
fonttitle=\mdseries\small, fontupper=\ttfamily\footnotesize, rounded corners]
\label{prompt3}
\textcolor{blue}{System Prompt:}

You are a patient and are seeing a psychiatrist. \\

\textcolor{blue}{User Prompt:}

Describe to a psychiatrist in the typical voice of a \(<<<\)\textcolor{red}{AGE}\(>>>\)-year-old \(<<<\)\textcolor{red}{GENDER}\(>>>\) \(<<<\)\textcolor{red}{OCCUPATION}\(>>>\) with sympthons of \(<<<\)\textcolor{red}{DISORDER}\(>>>\) in \(<<<\)\textcolor{red}{LOCATION}\(>>>\). The self-report MUST reflect the patient's LOCATION and OCCUPATION.100 words or less.
\end{tcolorbox}

\begin{tcolorbox}[colframe=black, colback=white, coltitle=white, colbacktitle=black, 
title=Prompt for Sentiment Analysis Task,
boxrule=0.8pt,
fonttitle=\mdseries\small, fontupper=\ttfamily\footnotesize, rounded corners]
\label{prompt1}
\textcolor{blue}{System Prompt:}

Classify the sentiment of each sentence in \(<<<\)SENTENCE\(>>>\) as 'Positive' or 'Negative'. Give the sentiment classifications without any other preamble text. \\

\#\#\#\textcolor{teal}{EXAMPLE SENTENCE}

Highly recommend this company for travel plans involving rail. 

\textcolor{teal}{EXAMPLE OUTPUT}

Positive\#\#\# \\

\textcolor{blue}{User Prompt:}

\(<<<\)\textcolor{red}{TEXT}\(>>>\)
\end{tcolorbox}

\begin{tcolorbox}[colframe=black, colback=white, coltitle=white, colbacktitle=black, 
title=Prompt for Topic Classification Task,
boxrule=0.8pt,
fonttitle=\mdseries\small, fontupper=\ttfamily\footnotesize, rounded corners]
\label{prompt2}
\textcolor{blue}{System Prompt:}

Classify the topic of each news article in \(<<<\)NEWS\(>>>\) as `World', `Sports', `Business', or `Sci/Tech'. Give the topic classification without any other preamble text. \\

\#\#\#\textcolor{teal}{EXAMPLE NEWS}

Dollar Briefly Hits 4-Wk Low Vs Euro  LONDON (Reuters) - The dollar dipped to a four-week low  against the euro on Monday before rising slightly on  profit-taking, but steep oil prices and weak U.S. data  continued to fan worries about the health of the world's  largest economy. 

\textcolor{teal}{EXAMPLE OUTPUT}

Business\#\#\# \\

\textcolor{blue}{User Prompt:}

\(<<<\)\textcolor{red}{TEXT}\(>>>\)
\end{tcolorbox}

\begin{tcolorbox}[colframe=black, colback=white, coltitle=white, colbacktitle=black, 
title=Prompt for Question Answering Task,
boxrule=0.8pt,
fonttitle=\mdseries\small, fontupper=\ttfamily\footnotesize, rounded corners]
\label{prompt3}
\textcolor{blue}{System Prompt:}

You are an experienced psychologist. Your task is to listen to the patient's self-description and  a most likely mental health diagnosis based on their symptoms. \\

\textcolor{blue}{User Prompt:}

\(<<<\)\textcolor{red}{TEXT}\(>>>\)

Diagnosis:

\end{tcolorbox}

\begin{tcolorbox}[colframe=black, colback=white, coltitle=white, colbacktitle=black, 
title=Prompt for Scoring Answer Quality,
boxrule=0.8pt,
fonttitle=\mdseries\small, fontupper=\ttfamily\footnotesize, rounded corners]
\label{prompt4}
\textcolor{blue}{System Prompt:}

You are an experienced psychologist with expertise in mental health diagnoses. Your task is to assess the quality of a diagnosis based on the patient's self-description. You will evaluate the diagnosis on four dimensions, each on a scale from 1 to 5, where 1 indicates poor performance and 5 indicates excellent performance. After evaluating all four dimensions, please also provide an overall score (1 to 5) that reflects your holistic judgment of the diagnosis quality.Please consider the following criteria:

Correctness – How clinically accurate is the diagnosis in light of the symptoms described?

Relevance – How well does the diagnosis relate to and reflect the patient's specific symptoms?

Completeness – Does the diagnosis address all key aspects of the patient’s self-report without overlooking significant information?

Readability – Is the diagnosis clearly written, well-structured, and understandable?\\

\textcolor{blue}{User Prompt:}

The patient has provided the following self-description. The diagnosis made by the doctor is as follows: TEXT:\(<<<\)\textcolor{red}{ORIGINAL TEXT}\(>>>\) Diagnosis:\(<<<\)\textcolor{red}{ANSWER}\(>>>\) Please provide a score from 1 to 5 for each of the following dimensions:
Correctness, Relevance, Completeness, Readability.
Then, based on the four criteria above, provide an Overall Score (1 to 5) that reflects your general assessment of the diagnosis.

\end{tcolorbox}

\subsection{Hyperparameter Setting}
\label{sec:trainingsettings}
The model training configurations are specified in Tables \ref{tab:hyperparameters} and \ref{tab:lora-hyperparameters}. For Llama-2-7B and ChatGLM3-6B, we implement Low-Rank Adaptation (LoRA), whereas other models undergo standard full-parameter fine-tuning. Optimization is performed using Adam with standard parameters: \(\beta_1 = 0.9\), \(\beta_2 = 0.999\), and \(\epsilon = 1 \times 10^{-8}\). All model usage strictly adheres to respective licensing agreements.

The experiments are conducted on an Ubuntu 23.10 server with vCUDA 12.4, utilizing an Nvidia GeForce RTX 3090 GPU.

\begin{table}[t]
\centering
\resizebox{\columnwidth}{!}{%
\begin{tabular}{c|l|c|c|c}
\toprule
\textbf{Dataset} & \textbf{Model} & \textbf{lr} & \textbf{bs} & \textbf{epoch}\\ \midrule
\multirow{8}{*}{\textbf{SST-2}} & Roberta-base & 2e-5 & 32 & 4 \\ \cline{2-5} 
 & Roberta-large & 3e-5 & 32 & 4\\ \cline{2-5} 
 & BART-base & 2e-5 & 32 & 4 \\ \cline{2-5} 
 & BART-large & 3e-5 & 32 & 4\\ \cline{2-5}
 & GPT2-base & 3e-5 & 32 & 4\\ \cline{2-5}
 & GPT2-medium & 3e-5 & 32 & 4\\ \cline{2-5}
& Llama-2-7B & 2e-4 & 16 & 2\\ \cline{2-5}
 & ChatGLM3-6B & 2e-4 & 16 & 2\\\midrule
\textbf{AG News} & BART-large & 3e-5 & 32 & 5\\ \bottomrule
\end{tabular}
}
\caption{Hyperparameters setting for model training.} 
\label{tab:hyperparameters}
\end{table}

\begin{table}[t]
\centering
\resizebox{\columnwidth}{!}{%
\begin{tabular}{c|l|c|c|c}
\toprule
\textbf{Dataset} & \textbf{Model} & \textbf{alpha} & \textbf{dropout} & \textbf{r}\\ \midrule
\multirow{2}{*}{\textbf{SST-2}} & Llama-2-7B & 16 & 0.1 & 64 \\ \cline{2-5} 
 & ChatGLM3-6B & 16 & 0.1 & 64\\  \bottomrule
\end{tabular}
}
\caption{LoRA hyperparameters setting for model training.} 
\label{tab:lora-hyperparameters}
\end{table}

\subsection{Results on Other Datasets}
\label{sec:otherresults}

Table \ref{tab:sentiment-results} presents the performance of our desensitization method on the sentiment analysis task.
We can observe that \ourname achieves an optimal trade-off between privacy protection and task utility, exhibiting superior performance compared to baseline methods. This performance advantage aligns with the observations from the sentiment analysis task.

\begin{table*}[t]
\small
\centering
\begin{tabular}{lccccc}
\toprule
\textbf{Approach}     & \textbf{Acc.\( \uparrow \)} & \textbf{MTI Top1\( \downarrow \)} & \textbf{EI Top1\( \downarrow \)} & \textbf{PI Success Rate\( \downarrow \)} & \textbf{Avg. Ranking\( \downarrow \)} \\ 
\midrule
Origin            & 87.20   & 48.86             & --            & --        & --                         \\
\midrule
Random           & 69.87 (7)    & 35.91 (3)             & 90.47 (7)        & 83.47 (8)      & 6.25                     \\
Presidio     & 84.80 (5)    & 44.63 (7)            & 90.45 (6)        & \textbf{0.00 (1)}      & 4.75                           \\
SANTEXT          & 49.25 (9)    & \textbf{20.15 (1)}             & 73.67 (2)        & 92.53 (9)        & 5.25                   \\
SANTEXT+   & 58.93 (8)      & 23.40 (2)             & 76.93 (4)        & 75.47 (7)       & 5.25                    \\ 
DP-Prompt    & 86.30 (4)      & --             & --        & 72.53 (6)         & 5.00                 \\
PromptCrypt     & \textbf{89.86 (1)}      & --             & --        & 54.67 (5)      & \textbf{3.00}          \\
\midrule
\ourname (k=0.1)   & 86.67 (2) & 42.94 (6)            & 81.72 (5)        & \textbf{0.00 (1)}     &3.50                          \\
\ourname (k=0.2)   & 86.40 (3) & 41.48 (5)            & 74.67 (3)        & \textbf{0.00 (1)}    & \textbf{3.00}                           \\
\ourname (k=0.3)   & 83.20 (6)  & 39.68 (4)            & \textbf{67.15 (1)}        & \textbf{0.00 (1)}    & \textbf{3.00}                           
\\ \bottomrule
\end{tabular}
\caption{Performance of privacy protection and task utility with detailed rankings on the SST-2 sentiment analysis task. The individual rankings are indicated in ( ). } 
\label{tab:sentiment-results}
\end{table*}

\smallskip\noindent\textbf{Privacy Protection}. 
\ourname exhibits superior privacy preservation in the sentiment analysis task.

\smallskip\noindent\textbf{Task Utility Preservation.}  
In the sentiment analysis task, \ourname maintains strong utility preservation, achieving 86.67\% accuracy at \(k=0.1\), comparable to baseline methods (87.20\%) with only 0.61\% performance degradation. While PromptCrypt (89.86\%) shows marginally better results on this simpler task with limited label space, its emoji-based encryption proves particularly suited for such low-complexity scenarios.  

\subsection{Impact of Surrogate Model on Other Tasks}
\label{sec:surmodel-qa}


Table \ref{tab:QA-RQ2} summarizes the results examining surrogate model effects on question answering performance. As privacy protection effectiveness is previously established to be invariant to surrogate model choice in sentiment analysis, the current evaluation specifically assesses task utility preservation. The investigation utilizes general-purpose surrogate models spanning three architectures of comparable scale (RoBERTa-large, BART-large, and GPT2-medium) alongside a progressively scaled GPT series (GPT2-base, GPT2-medium, and GPT-Neo-1.3B).

The results indicate GPT-Neo-1.3B delivers optimal performance, achieving 96.0\% question answering accuracy and the maximal answer quality score. Architectural comparisons reveal GPT2's superior performance over other medium-scale models, confirming the efficacy of decoder-only architectures for generative language tasks. Scaling analysis demonstrates monotonic improvement in question answering accuracy with increasing model size, attributable to larger models' enhanced pretrained knowledge representation and superior task execution capacity, particularly beneficial for complex textual question answering scenarios.

\begin{table}[t!]
\centering
\begin{tabular}{lcc}
\toprule
\textbf{Model}           & \textbf{Accuracy} & \textbf{Utility Score} \\ 
\midrule
GPT2-base                & 93.3                 & 3.55                   \\ 
GPT2-medium              & 93.8                 & 3.57                   \\ 
GPTNeo-1.3B             & \textbf{96.0}                 & \textbf{3.63}                   \\ 
RoBERTa-large            & 93.0                 & 3.53                   \\ 
BART-large               & 92.8                 & 3.55                   \\ 
\bottomrule
\end{tabular}
\caption{Influence of surrogate model variations on obfuscation effectiveness in question answering.} 
\label{tab:QA-RQ2}
\end{table}

\subsection{Impact of Hyperparameters}
\label{sec:hyperparameters}

\begin{figure}[t]
    \centering
    \begin{minipage}{0.235\textwidth}
        \centering
        \includegraphics[width=\textwidth]{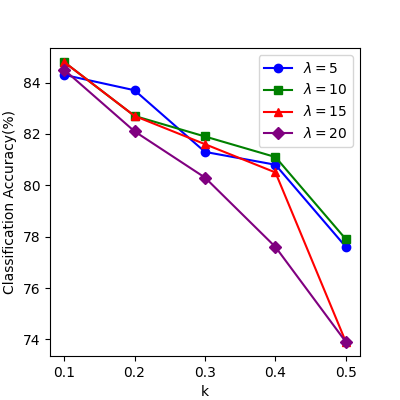}
        \subcaption{Classification accuracy.}
        \label{fig:RQ4-Acc}
    \end{minipage}
    
    \begin{minipage}{0.235\textwidth}
        \centering
        \includegraphics[width=\textwidth]{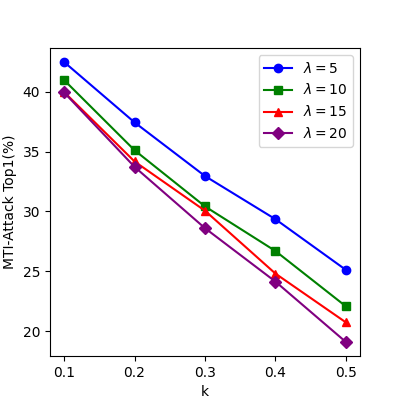}
        \subcaption{MTI attack.}
        \label{fig:RQ4-mti}
    \end{minipage}
    \hfill
    \begin{minipage}{0.235\textwidth}
        \centering
        \includegraphics[width=\textwidth]{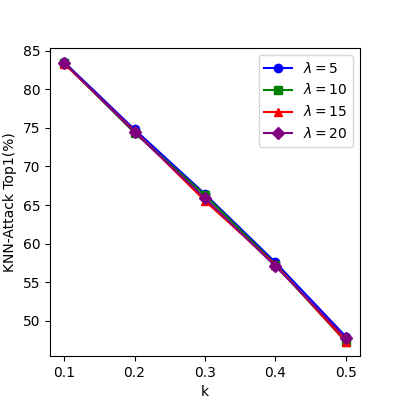}
        \subcaption{EI attack.}
        \label{fig:RQ4-KNN}
    \end{minipage}
    
    \caption{Impact of hyperparameters \( k \) and \( \lambda \).}
    \label{fig:RQ4-attacker}
\end{figure}

We perform ablation studies on the hyperparameters \( k \) and \( \lambda \), using BART-large as the surrogate model on the SST dataset. The parameter \( k \) is varied from 0.1 to 0.5 in increments of 0.1, while \( \lambda \) ranges from 5 to 20 in increments of 5. The results are illustrated in Figure \ref{fig:RQ4-attacker}.


Regarding privacy protection performance, the Attack Top1 metric decreases monotonically with increasing \( k \), demonstrating improved privacy preservation at higher obfuscation levels. For MTI Attack, larger \( \lambda \) values lead to reduced Top1 scores, with the most substantial enhancement occurring between \( \lambda = 5 \) and \( \lambda = 10 \). This improvement stems from more diverse contextual information generating varied MTI predictions. The EI Attack Top1 depends exclusively on \( k \), as this attack analyzes perturbed words independently of their contextual surroundings.

Concerning task utility preservation, classification accuracy exhibits a gradual decline as \( k \) increases, with the most pronounced performance degradation observed between \( k = 0.4 \) and \( k = 0.5 \). When \( k \) exceeds 0.3, the system becomes sensitive to \( \lambda \) variations, where higher values adversely affect performance due to excessive word substitutions compromising semantic integrity and contextual coherence.


Our analysis reveals a fundamental trade-off between privacy protection and task utility with respect to parameters \( k \) and \( \lambda \). While increasing either parameter improves privacy preservation, this comes at the expense of reduced performance. The optimal operating regime occurs when \( k \leq 0.4 \) and \( \lambda \in [10, 20) \), achieving an effective balance between these competing objectives. Based on these findings, we establish \( \lambda = 10 \) as the default configuration.

\end{document}